%% file: acl_latex.tex
\pdfoutput=1

\documentclass[11pt]{article}

\usepackage[final]{acl}

\usepackage{times}
\usepackage{latexsym}

\usepackage[T1]{fontenc}

\usepackage[utf8]{inputenc}

\usepackage{microtype}

\usepackage{inconsolata}

\usepackage{graphicx}

\usepackage{hyperref}
\usepackage{booktabs}
\usepackage{amssymb}
\usepackage{makecell}

\newcommand{\ie}{\textit{i}.\textit{e}.}

\usepackage{pifont}
\newcommand{\cmark}{\ding{51}}
\newcommand{\xmark}{\ding{55}}
\usepackage{arydshln}
\usepackage{wrapfig}
\usepackage{multirow}
\usepackage{amsmath}
\usepackage{verbatim}

\title{Dub-S2ST: Textless Speech-to-Speech Translation for Seamless Dubbing}

\author{Jeongsoo Choi$^*$ \quad Jaehun Kim$^*$ \quad Joon Son Chung\\
Korea Advanced Institute of Science and Technology\\
\small{\texttt{\{jeongsoo.choi, kjaehun, joonson\}@kaist.ac.kr}}}

\begin{document}

\maketitle

\begingroup
\def\thefootnote{}\footnotetext{$^*$Equal contribution.}
\endgroup

\begin{abstract}
This paper introduces a cross-lingual dubbing system that translates speech from one language to another while preserving key characteristics such as duration, speaker identity, and speaking speed. Despite the strong translation quality of existing speech translation approaches, they often overlook the transfer of speech patterns, leading to mismatches with source speech and limiting their suitability for dubbing applications. To address this, we propose a discrete diffusion-based speech-to-unit translation model with explicit duration control, enabling time-aligned translation. We then synthesize speech based on the translated units and source speaker's identity using a conditional flow matching model. Additionally, we introduce a unit-based speed adaptation mechanism that guides the translation model to produce speech at a rate consistent with the source, without relying on any text. Extensive experiments demonstrate that our framework generates natural and fluent translations that align with the original speech's duration and speaking pace, while achieving competitive translation performance. The code is available at \url{https://github.com/kaistmm/Dub-S2ST}.
\end{abstract}

\section{Introduction}
Recent advancements in translation systems and speech technologies have enabled a vast amount of multimedia content to support multiple languages through automated speech dubbing. Cross-lingual dubbing~\cite{federico2020speech, wu2023videodubber}, which replaces speech audio of one language with that of another, allows global audiences to consume content in their native languages. While this significantly reduces language barriers, ensuring effective dubbing requires meeting several specific criteria: maintaining duration, speaker identity, and speaking speed~\cite{brannon2023dubbing}.

Traditional dubbing systems typically employ a cascade of Automated Speech Recognition (ASR)~\cite{amodei2016deep, baevski2020wav2vec, radford2023robust}, Neural Machine Translation (NMT)~\cite{johnson2017google, stahlberg2020neural, fan2021beyond, costa2022no}, and Text-to-Speech (TTS)~\cite{wang2017tacotron, ren2021fastspeech, wang2023neural, tan2024naturalspeech} modules. Although these cascaded systems demonstrate promising translation quality, they inherently lose critical speech-related information, such as speaker identity and prosody, due to the intermediate text representations~\cite{swiatkowski2023cross}. Moreover, because text lacks precise duration information, these systems struggle to accurately match the duration and speaking pace of the original speech~\cite{sahipjohn2024dubwise}. As a result, even with post-processing, the output often remains misaligned or unnatural, limiting their effectiveness for real-world dubbing~\cite{effendi2022duration}.

To address these limitations, Speech-to-Speech Translation (S2ST) systems~\cite{jia2019translatotron, jia2022translatotron, barrault2025seamless} have emerged and evolved into textless S2ST~\cite{lee2021textless, li2023textless, kim2024textless}, aiming to translate speech directly without intermediate text. While recent approaches have achieved translation quality comparable to cascaded systems, most of them lack the capability to control speech duration during translation. As a result, the output typically requires post-processing like manual stretching or contracting to match the original speech duration for dubbing purposes. However, this process can degrade speech quality and lead to unnatural prosody and speaking pace.

A key challenge underlying this issue lies in the limitations of existing training datasets. High-quality dubbing demands not only accurate translation but also faithful preservation of the source speech’s voice characteristics and speaking speed. However, it is inherently difficult to construct large-scale datasets containing the same speaker uttering aligned content in multiple languages. Consequently, most S2ST datasets rely on synthesized target speech~\cite{jia2022cvss} or web-crawled data~\cite{duquenne2022speechmatrix, barrault2025seamless}, which prioritize linguistic fidelity over speech-related information consistency. These datasets tend to exhibit discrepancies in speaker identity and speaking speed, making it challenging for models to learn how to preserve such attributes. Despite progress in translation accuracy, existing S2ST models trained on these datasets remain suboptimal for seamless dubbing.

In response, we propose Dub-S2ST, a novel textless S2ST framework specifically designed for dubbing applications, effectively leveraging existing datasets. To mitigate the ambiguity caused by discrepancies between source and target speech, we eliminate variations in the target that deviate from the paired source. Specifically, we first convert continuous speech to discrete units that retain rich semantic features and minimal acoustic variations~\cite{lakhotia2021generative, lee2021textless}. We then apply our unit-based speed adaptation strategy to adjust the target's speaking rate to the source.
Using the processed data, we develop a speech-to-unit translation model trained with a discrete diffusion objective~\cite{austin2021structured}. We also incorporate Diffusion Transformer~\cite{peebles2023scalable}, which allows the model to accurately predict speech units conditioned on diffusion timestep. Moreover, our model inherently supports duration control by using predetermined lengths based on the duration of source speech.
Finally, we incorporate a conditional flow matching (CFM)~\cite{lipman2023flow, mehta2024matcha}-based synthesizer that generates high-quality speech conditioned on the translated speech units and the original source speech, closely resembling the original speaker's identity.

Through extensive evaluation, our proposed framework demonstrates superior preservation of duration, speaker identity, and speaking speed, while maintaining competitive translation accuracy. Ablation studies further validate the effectiveness of each component in improving dubbing quality. 
To the best of our knowledge, Dub-S2ST is the first textless S2ST framework tailored for seamless automatic dubbing that preserves both speaker identity and speaking speed.

\begin{figure*}[t]
	\centering
	\centerline{\includegraphics[width=\linewidth]{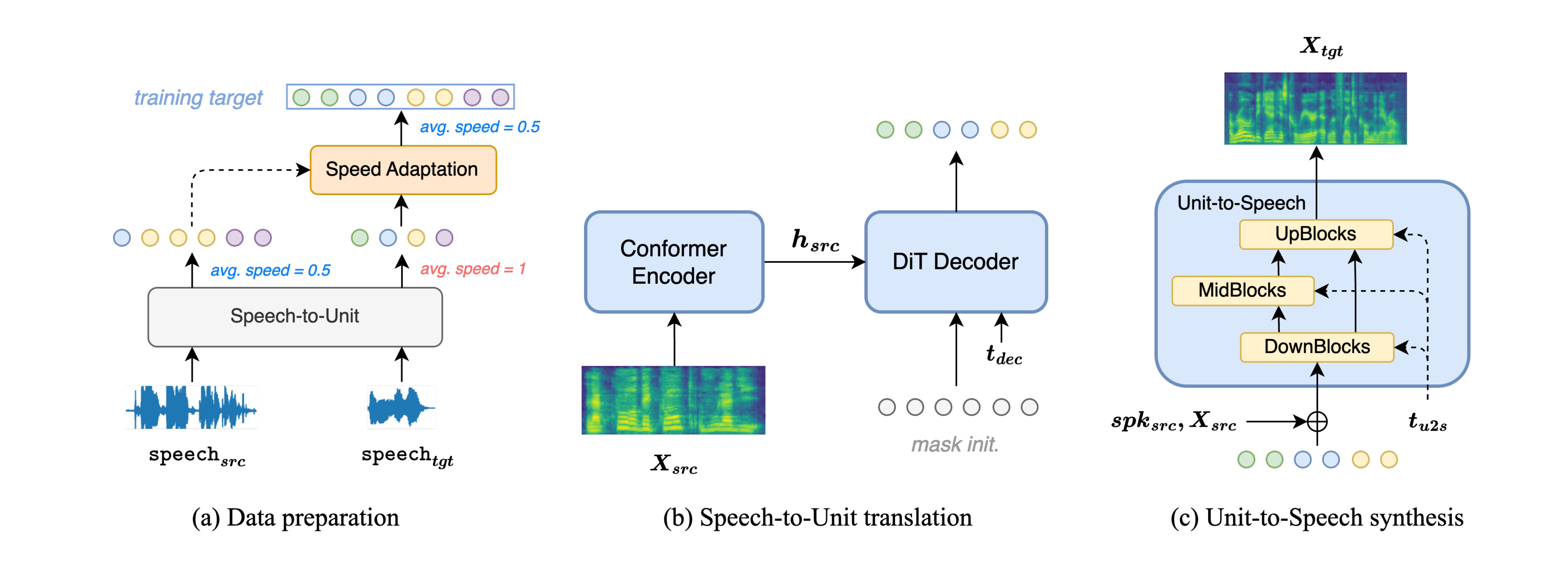}}
        \vspace{-0.1cm}
        \caption{
        Dub-S2ST framework. (a) \textit{avg. speed} indicates average unit speed calculated in unit-based speed adaptation. (b) $h_{src}$ indicates the encoded source speech features from the encoder, and $t_{dec}$ is the timestep information used to train discrete-diffusion decoder. (c) $spk_{src}, X_{src}$ represent speaker embedding and mel-spectrogram from the source speech as conditions, respectively. $t_{u2s}$ is the timestep information used to train unit-to-speech synthesizer.
        }
        \vspace{-0.1cm}
	\label{fig:main}
\end{figure*}

\section{Related Works}
\subsection{Cross-lingual Dubbing}
Dubbing is a post-production process in which the original spoken dialogue in multimedia content is replaced with speech in another language, while preserving the temporal alignment and naturalness of the original speech~\cite{orero2004topics}. Early automatic dubbing systems typically adopt cascaded S2ST architecture, combining ASR, NMT, and TTS~\cite{dureja2015speech}. While maintaining this cascaded pipeline, recent efforts have focused on enhancing each component. Some approaches focus on improving prosodic alignment, aiming to synchronize the prosody of the generated speech with the original~\cite{federico2020evaluating}. Others leverage existing TTS models by modifying the duration module to generate speech that matches the original duration~\cite{effendi2022duration}. Despite these advancements, relying on text as intermediate representation inherently limits temporal flexibility, highlighting the need for textless approaches that better preserve the naturalness of the source speech.

\subsection{Speech-to-Speech Translation (S2ST)}
Speech-to-Speech Translation (S2ST) aims to convert source speech into a target language while preserving linguistic content. Early systems adopted a cascaded architecture, integrating ASR, MT, and TTS modules~\cite{federico2020evaluating, lakew2022isometric}. To mitigate error propagation and latency issues inherent in cascaded systems, direct S2ST approaches have been introduced~\cite{jia2019translatotron, jia2022translatotron}, demonstrating the feasibility of end-to-end speech translation. To further eliminate reliance on intermediate text, textless S2ST methods have emerged. S2UT~\cite{lee2021textless} proposes an autoregressive (AR) translation model that predicts deduplicated discrete speech units~\cite{lakhotia2021generative}, which are then used to synthesize target speech. UTUT~\cite{kim2024textless} extends this framework to support many-to-many language translation. Subsequent works like TranSpeech~\cite{huang2023transpeech} explore non-autoregressive (NAR) models to achieve faster decoding compared to AR models, and DiffNorm~\cite{tan2024diffnorm} enhances TranSpeech by introducing techniques to normalize acoustic variations in the target speech. More recently, CTC-S2UT~\cite{fang2024ctc} incorporates CTC~\cite{graves2006connectionist}-based unit reduction to improve translation performance. Unlike previous approaches, we leverage a NAR-based model without deduplicating speech units. This design preserves temporal information and enables explicit control over output duration, making our framework naturally suitable for dubbing scenarios.

\subsection{S2ST Datasets}
Recent advancements in S2ST have been facilitated by the development of specialized datasets. VoxPopuli~\cite{wang2021voxpopuli} is a multilingual corpus containing aligned speech pairs derived from simultaneous interpretation by human interpreters at European Parliament events. This dataset provides realistic translation pairs, however, it is limited in terms of domain coverage and scalability. A common approach to handle these issues is synthesizing target speech from translated text~\cite{jia2019translatotron, jia2022translatotron}. For instance, CVSS~\cite{jia2022cvss} is built by converting the text from the speech-to-text translation corpus CoVoST2~\cite{jia2022cvss} into speech using a pretrained TTS model. Another approach is based on data mining. SeamlessAlign~\cite{barrault2025seamless} leverages web-crawled multimodal translation data and employs a unified speech-text similarity model~\cite{duquenne2023sonar} to effectively pair speech segments, resulting in approximately 29,000 hours of S2ST data. However, these datasets still lack consistency in speech attributes between source and target. Our goal is to develop a framework that can generate dubbed speech that closely follows the source, even under such conditions.

\section{Method}
In this section, we explain the data preparation, model architecture and training objective, and final synthesis of the translated speech. The overall architecture of our model is illustrated in  Fig~\ref{fig:main}.

\subsection{Data Preparation}
\noindent\textbf{Speech Unit Extraction.}
The choice of target speech representation plays a critical role in determining the quality and accuracy of S2ST. While continuous features allow for straightforward generation, they are often susceptible to noise and mispronunciations, resulting in reduced intelligibility and naturalness. In contrast, discrete features leverage pretrained speech Self-Supervised Learning (SSL) models followed by quantization~\cite{hsu2021hubert, lee2021textless}. Empowered by large-scale SSL training and discretization, this approach effectively captures phonetic information while minimizing speaker-dependent attributes such as timbre and pitch~\cite{lakhotia2021generative}.

To focus explicitly on the linguistic components of speech during translation, the proposed method adopts units extracted using mHuBERT~\cite{lee2022direct}, followed by k-means clustering for quantization. This cascaded unit extraction process facilitates robust representation of linguistic content while suppressing paralinguistic variations.

\noindent\textbf{Unit-Based Speed Adaptation.}
Speaking speed is an essential factor in dubbing, as perceptual quality significantly degrades when the speed of translated speech deviates from the original~\cite{orero2004topics}. While signal processing techniques can be applied to adjust speech duration, they often compromise naturalness and intelligibility. Another strategy is to guide models using syllable- or phoneme-based speed metrics, since these provide an indication of speaking pace across languages~\cite{barrault2025seamless}; however, such methods are inapplicable in textless systems.

To address these limitations, we propose a unit-based speed adaptation method that adjusts the repetition of speech units based on the speaking speed ratio between source and target speech. This method is inspired by unit deduplication (\ie, repetition removal) used in recent S2ST models~\cite{lee2021textless, huang2023transpeech, fang2024ctc} to preserve synthesis quality while minimizing redundancy. We hypothesize that the reduced sequence $\hat{L}$ captures a distinct set of pronunciations, and the ratio against original length $L$ serves as an implicit estimate of speaking speed $r=\frac{\hat{L}}{L}$.

After extracting speech unit sequences from both source and target, we adjust the target sequence by applying the speed ratio $\frac{r_{src}}{r_{tgt}}$, modifying the number of unit repetitions to align the target speed with the source. The speed is normalized by the average speed of each language to mitigate cross-linguistic differences. This adaptation preserves linguistic content while controlling speaking rate. Note that, as depicted in Figure~\ref{fig:main}, the speed adaptation method does not force the target speech units to have the same length as the source. It rather changes the rate of repeating units, so that it only alters the implicit speaking pace. Training the model on speed-aligned sequences enables it to generate translated speech that naturally and consistently mirrors the source speaking speed.

\subsection{Speech-to-Unit Translation}
Accurate dubbing requires the translated speech to match the duration of the source utterance. To fulfill this requirement, we propose a speech-to-unit translation model whose decoder is implemented as a discrete diffusion~\cite{austin2021structured} generator conditioned on the source speech. The decoder leverages Diffusion Transformer (DiT)~\cite{peebles2023scalable} layers that operate on variable-length sequences, with additional cross-attention to integrate source speech features throughout the generation. During inference, the decoder takes a fully masked unit sequence with length identical to that of the source speech as its initial input and iteratively transforms it into speech units.

The training is conducted by first masking units based on a mask schedule $\gamma(t_{dec})$, where both $t_{dec}$ and $\gamma(t_{dec})$ range from $(0, 1)$. Each unit is independently masked with probability $\gamma(t_{dec})$ and remains unmasked with probability $1-\gamma(t_{dec})$, and the decoder predicts the original target units from the partially masked sequence. The decoder is optimized using the following cross-entropy loss:
\begin{equation}
    \mathcal{L} = -\mathbb{E} \left[ \frac{1}{|M|} \sum_{i \in M} \log p_{\theta}(x^i_0 \mid x_t, h_{\text{src}}, t_{dec}) \right]
\end{equation}
Here, $M$ is the set of indices corresponding to masked units, $x^i_0$ is the ground-truth speech unit, $x_t$ is the partially masked input sequence at timestep $t_{dec}$, and $h_{src}$ is the encoded representation from the Conformer encoder. During training, the decoder predicts speech units from partially masked target sequence. During inference, it receives a fully masked sequence whose length matches that of the source speech units, thereby generating translated speech aligned in length with the source. The loss is applied only to the masked units, and we further investigate the impact of this in Section~\ref{sec:result}.

To enhance the decoder's reliance on source speech representations, we initialize the encoder from a pretrained autoregressive speech-to-unit translation model~\cite{lee2022direct} and fine-tune the entire weights. Additionally, we apply label-smoothing~\cite{szegedy2016rethinking} with a factor empirically set to 0.01 to improve generalization.

\subsection{Unit-to-Speech Synthesis}
A key factor in the perceptual quality of dubbing is the similarity between the translated speech and the original utterance. To convert semantic units to speech while preserving the acoustic characteristics of the source speech, we employ a unit-to-speech synthesizer based on Optimal Transport Conditional Flow Matching (OT-CFM). The synthesizer is implemented as a U-Net~\cite{rombach2022high} architecture, where each layer is a block comprising Convolutional and Transformer layers. Downsampling and upsampling are performed in the latent space, allowing the model to efficiently reconstruct fine-grained temporal structure.

We train the model using the OT-CFM objective, which defines a time-dependent vector field that transports a sample from a simple prior distribution \( x_0 \sim \mathcal{N}(0, I) \) to a data sample \( x_1 \sim q \) via linear displacement interpolation:
\begin{equation}
    \varphi_t = \left(1 - (1 - \sigma_{\min})t_{u2s}\right)x_0 + t_{u2s}x_1
\end{equation}
The decoder is trained to minimize the difference between the predicted and target velocities:
\begin{equation}
    \mathcal{L} = \mathbb{E}_{t_{u2s}, x_0, x_1} \left\|
    u_t(\varphi_t \mid x_1) - 
    v_\theta(\varphi_t | t_{u2s}, \boldsymbol{c})
    \right\|^2
\end{equation}
where the ground-truth velocity is given by:
\begin{equation}
    u_t^{\text{OT}}(\varphi_t^{\text{OT}} \mid x_1) = x_1 - (1 - \sigma_{\min})x_0
\end{equation}
The condition $\boldsymbol{c}$ consists of the unit embedding sequence from source and target speech, the source speaker embedding $spk_{\text{src}}$ extracted with a pretrained speaker verification model~\cite{wang2023cam++}, and the source mel-spectrogram $X_{\text{src}}$. The mel-spectrogram is padded to match the length of the unit sequence, while the speaker embedding is repeated accordingly. These features are concatenated channel-wise to enable in-context learning: the sampled prior is transformed into a mel-spectrogram conditioned on both speaker identity and prosodic information from the source speech.

Although directly training the module on S2ST data is possible, such datasets often contain noise and reverberation, which can impair synthesis quality. To address this, we initialize the model from a TTS model~\cite{du2024cosyvoice} trained on multilingual corpus, and fine-tune it with necessary adaptations for our semantic unit input. This approach allows for robust zero-shot synthesis across diverse speakers and languages. The generated mel-spectrogram is then converted to audible waveform via a pretrained HiFi-GAN~\cite{kong2020hifi}.

\section{Experiment}
\subsection{Dataset}
\noindent\textbf{CVSS-C}~\cite{jia2022cvss} is a widely-used dataset for speech-to-speech translation, consisting of 21 languages to English translations where English speech is generated by a single-speaker TTS model. The proposed method is trained and evaluated with French-English (fr-en) subset, due to the abundance of samples compared to other language pairs. The fr-en subset contains 207,364 (train) / 14,759 (dev) / 14,759 (test) pairs of source and target speech samples, totaling 264 hours. The experiment utilizes train and dev split for training and validation, and test split for final evaluation.

\subsection{Implementation Details}
\noindent\textbf{Preprocessing.}
Audio samples are resampled to 16kHz and preprocessed with Voice Activity Detection (VAD) tool\footnote{\url{https://github.com/snakers4/silero-vad}} to remove unnecessary silence and paralinguistic information at the beginning and end.
The samples are then processed with pretrained mHuBERT~\cite{lee2021textless} and k-means clustering model to obtain discrete units\footnote{\url{https://github.com/facebookresearch/fairseq/blob/main/examples/speech_to_speech/docs/textless_s2st_real_data.md}}.

\noindent\textbf{Architecture.}
To maintain consistency with prior works on speech-to-unit translation models, we design our model with 12 Conformer encoder layers and 6 DiT layers, totaling 61M parameters. The unit-to-speech synthesizer consists of 4 Down, Mid, and UpBlock, where each is a cascade of 1D Convolution and Transformer layer.

\noindent\textbf{Training.}
The speech-to-unit translation model is trained with a total batch size of 3,200 seconds for 100k updates.
We use 8 RTX A5000 GPUs for training, and the total training takes approximately 10 hours.
We optimize the model with the AdamW~\cite{loshchilov2019decoupled} optimizer and applied a dropout rate of 0.3. The learning rate is warmed up for the first 10k steps to a peak of $1 \times 10^{-3}$, and then decayed using an inverse square root schedule. We implement our approach using Fairseq~\cite{ott2019fairseq}.
The unit-to-speech module is initialized from CosyVoice-300M~\cite{du2024cosyvoice} and fine-tuned with fixed learning rate of $1 \times 10^{-4}$ for 200k steps using LRS3~\cite{afouras2018lrs3} dataset, an English multi-speaker corpus.

\subsection{Evaluation Metrics}
\noindent\textbf{ASR-BLEU}~\cite{lee2022direct} is a widely adopted metric for assessing S2ST quality. It measures translation quality by transcribing the generated speech using a pretrained ASR model~\cite{baevski2020wav2vec} and comparing it with the ground-truth text to compute BLEU~\cite{post2018call} \footnote{\url{https://github.com/facebookresearch/fairseq/tree/ust/examples/speech_to_speech/asr_bleu}}.

\input{Table/main_translation}

\input{Table/main_duration}

\noindent\textbf{BLASER 2.0}~\cite{dale2024blaser} serves as an automatic measure for assessing semantic similarity between source and generated speech.
We adopt its reference-free variant, BLASER 2.0-QE\footnote{\mbox{\url{https://huggingface.co/facebook/blaser-2.0-qe}}}, which estimates translation quality without reference text unlike BLEU score.

\noindent\textbf{SIM} evaluates the speaker similarity between the generated and original speech. We use a pretrained speaker verification model~\cite{chen2022large}, based on WavLM-Large~\cite{chen2022wavlm}, to extract speaker embedding vectors and calculate cosine similarity between the two.

\noindent\textbf{DNSMOS} (Deep Noise Suppression Mean Opinion Score)~\cite{reddy2021dnsmos} is an automated perceptual speech quality assessment of generated speech\footnote{We use a model trained with ITU-T P.808: \url{https://github.com/microsoft/DNS-Challenge/tree/master/DNSMOS}}.
The quality is estimated with a score in the range of [1, 5] where the larger value indicates higher speech quality.

\noindent\textbf{Duration Compliance (DC)}~\cite{wu2023videodubber} calculates the portion of generated speech whose duration ratio with the source lies within certain range. This indicates how S2ST system preserves the duration when generating translated speech.

\noindent\textbf{Speed Compliance (SC)} captures a similar aspect to DC but is based on the ratio between the speed of speech, measured in syllables per second\footnote{\url{https://github.com/facebookresearch/seamless_communication/blob/main/docs/expressive}}. This metric reflects how closely the speed of the generated speech aligns with that of the source.

\section{Experimental Results}\label{sec:result}

\subsection{Quantitative Comparison}
\noindent\textbf{Translation Quality.}
Table~\ref{tab:translation} presents various evaluation results comparing the proposed method against existing baselines. Dub-S2ST-single, which employs the unit vocoder of S2UT for fair comparison with single-speaker approaches, outperforms all existing duration-controllable methods, achieving a BLEU score of 23.88. Disabling our proposed speed adaptation strategy leads to performance drops across all metrics, highlighting its effectiveness. A more detailed analysis of the speed adaptation is presented in the following section.

In addition, the last row in the table reports the performance of Dub-S2ST using our proposed multi-speaker unit-to-speech synthesizer. It shows superior speaker identity preservation, outperforming all baselines in terms of speaker similarity. Furthermore, Dub-S2ST achieves highly competitive ASR-BLEU and BLASER 2.0 scores compared to S2UT and CTC-S2UT, which lack duration control. This indicates that our model effectively captures the semantic information from the source speech and transfers it to the translated speech, while maintaining the duration and identity.

\input{Table/main_mos}

\noindent\textbf{Duration and Speed Analysis.}
We evaluate our model’s performance in duration control by measuring the generated speech duration relative to the source, as shown in Table~\ref{tab:duration}. In addition to standard S2ST baselines, we compare against methods that incorporate duration control: VideoDubber~\cite{wu2023videodubber}, which employs additional positional embeddings, and TransVIP~\cite{le2024transvip}, which introduces isochrony positional embedding based on voice activity information. While existing methods achieve reasonable compliance within a 40\% threshold, they struggle under a stricter 20\% constraint. In contrast, our model achieves exact duration matching through explicit length initialization in the decoder, yielding 100\% compliance.

The proposed model’s advantage becomes more evident in the speed analysis, where our model, both with and without speed adaptation, outperforms all baselines in speed compliance. This indicates that, when conditioned on the source duration, our model not only matches the duration but also the speaking pace of the source. The proposed speed adaptation further enhances this alignment. Overall, these results underscore the robustness of our approach in controlling speech duration and speed, both of which serve as critical factors in dubbing.

\begin{figure}[t]
	\centering
	\centerline{\includegraphics[width=\linewidth]{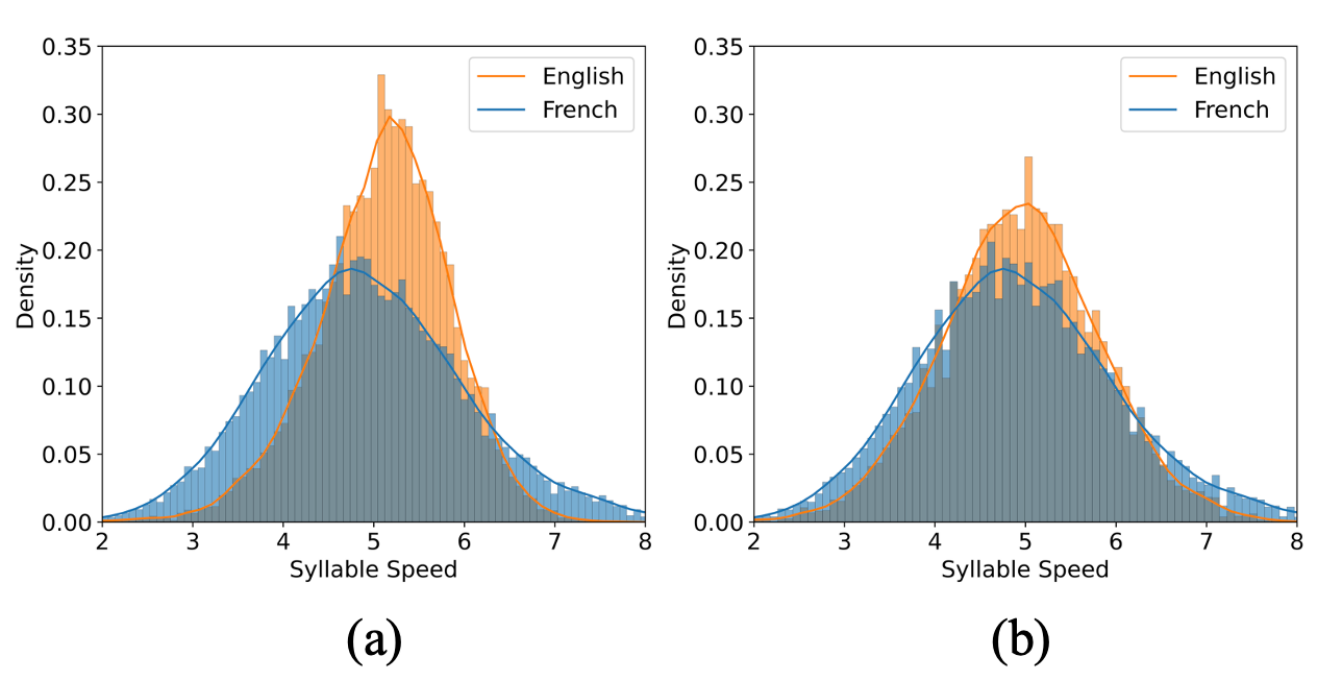}}
        \vspace{-0.2cm}
        \caption{Change in syllable speed distribution on the CVSS-C dataset (a) before and (b) after applying our speed adaptation.}
        \vspace{0.1cm}
	\label{fig:speed_adaptation}
\end{figure}

\input{Table/speed_adaptation}

\subsection{Human Evaluation}
We conducted a Mean Opinion Score (MOS) survey to assess the perceptual quality of the generated speech, as presented in Table~\ref{tab:mos}. 15 professional listeners rated samples on a scale from 1 (poor) to 5 (excellent). To ensure fair comparison, samples with durations differing from the source speech were manually adjusted to match the original.

The speed consistency evaluation confirms that our proposed speed adaptation method effectively aligns the speaking speed between source and target speech, producing translated speech at a pace closely matching the source. This implicit synchronization contributes to improving the translation consistency, since it helps the model generate sequences at a similar speed to the source when given its duration, minimizing unnecessary pauses and repetitive words. Notably, significant differences in speech naturalness emerged during the evaluation. Baseline samples required manual waveform adjustments, leading to considerable degradation in perceptual quality. Conversely, our proposed approach yielded accurate, time-aligned speech outputs, eliminating the need for post-processing adjustments and maintaining high naturalness suitable for dubbing applications.

\begin{figure}[t]
	\centering
	\centerline{\includegraphics[width=\linewidth]{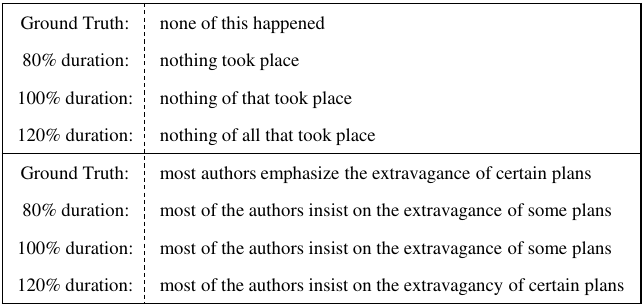}}
        \vspace{-0.1cm}
        \caption{Examples of ASR transcribed translation outputs of our model with varying target durations.}
        \vspace{0.1cm}
	\label{fig:duration_control}
\end{figure}

\input{Table/duration_control}

\subsection{Discussion}
\noindent\textbf{How well does unit speed reflect syllable speed?}
Syllable rate refers to the number of syllables spoken per second and is commonly used for measuring speaking speed. As a preliminary analysis, we measure the correlation between our unit-based speech rate estimation and syllable-based speed derived from transcripts. As shown in Table~\ref{tab:speed_adaptation}, the proposed metric strongly correlates with ground-truth syllable speed, validating its reliability.

\noindent\textbf{How effective of unit-based speed adaptation?}
Figure~\ref{fig:speed_adaptation} illustrates the effect of the proposed speed adaptation strategy. In Figure~\ref{fig:speed_adaptation} (a), the syllable speed distribution of English utterances deviates from that of French. However, after applying speed adaptation, the English distribution in Figure~\ref{fig:speed_adaptation} (b) closely aligns with the French distribution. This indicates that the proposed adaptation method effectively adjusts the speaking pace of the target.
 
\noindent\textbf{Can explicit duration control truly change the translated content?}
As illustrated in Figure~\ref{fig:duration_control}, our model responds to different duration prompts by producing semantically consistent outputs with nuanced variations in structure and phrasing. For example, “none of this happened” is rendered as “nothing took place” at 80\% duration and “nothing of all that took place” at 120\%, each expressing the same meaning without repetition or loss of information. Moreover, the relative number of characters of the translated outputs with different durations in Table~\ref{tab:duration_control} validates that the model flexibly generates semantics that fit the given duration, rather than forcing a fixed translation to fit varying lengths.

\input{Table/ablation_arch}

\input{Table/ablation_loss}

\subsection{Ablation Study}
\noindent\textbf{Model architecture and Masking schedule.}
To assess how different model architectures and masking schedules affect the speech-to-unit translation model of Dub-S2ST, we compare several design choices, as shown in Table~\ref{tab:arch}. The last row presents ASR-BLEU performances from Dub-S2ST-single without speed adaptation, evaluated across varying numbers of function evaluations (NFE). Replacing our DiT decoder with a standard Transformer decoder consistently degrades translation quality across all NFEs, indicating that incorporating diffusion timestep into the model benefits translation learning.
We also examine the effect of the masking schedule by comparing our linear schedule with a cosine schedule. The results show that the linear schedule yields overall better performance, suggesting its effectiveness without losing translation quality.
Based on the latency-performance trade-off, we choose the NFE of 64 for our evaluation.

\noindent\textbf{Loss computation.}
In Table~\ref{tab:loss}, we examine the impact of different loss computation strategies on our model's performance. Our findings indicate that computing loss on all units (\ie, predicting both masked and non-masked units) results in lower translation performance than masked prediction. This outcome is primarily due to the model's tendency to focus on easier predictions, leading to ineffective training. Calculating loss only on masked units, as implemented in our method, yields the best performance across all translation quality metrics. We also evaluated grouped masking strategy, where loss is not computed on masked units if at least one unit in the repeating group is unmasked, but it results in lower performance.

\input{Table/ablation_vocoder}

\noindent\textbf{Unit-to-speech module.}
To evaluate our unit-to-speech synthesizer’s ability to preserve speaker identity, we compare the performance with a zero-shot vocoder proposed in AV2AV~\cite{choi2024av2av}. Additionally, we applied a cross-lingual voice conversion model~\cite{du2024cosyvoice} to our single-speaker model and compare the results. As shown in Table~\ref{tab:vocoder}, our unit-to-speech model achieves the highest translation quality, even outperforming the single-speaker model. While the zero-shot vocoder maintains reasonable translation quality, it shows poor speaker similarity. On the other hand, using a separate voice conversion model shows a significant drop in ASR-BLEU. Based on qualitative analysis, we hypothesize that the process of voice conversion introduces oversmoothing of pronunciation that leads to loss in intelligibility and ultimately translation quality.

\section{Conclusion}
In this paper, we introduce an S2ST system suitable for dubbing applications. The proposed model generates translated speech with accurate content while preserving the duration and speaking speed of the source. This is achieved by the unique design that can generate speech with arbitrary duration, and speed adaptation that mitigates discrepancies between source and target speech. Extensive experiments with systematic ablations demonstrate that Dub-S2ST outperforms the existing baselines and verify its applicability to cross-lingual dubbing.

\section{Limitations}
While the proposed model offers an effective approach to cross-lingual dubbing, it is trained on speech recorded in controlled environments. This suggests that training on larger and more diverse datasets may be required for application to in-the-wild scenarios. Moreover, the proposed model is trained on a sentence-level speech corpus, and therefore a segmentation process is required before being utilized in other applications.


\section*{Acknowledgments}
This work was supported by Institute of Information \& communications Technology Planning \& Evaluation (IITP) grants funded by the Korean government (MSIT, RS-2025-02215122, Development and Demonstration of Lightweight AI Model for Smart Homes and RS-2022-II220989, Development of Artificial Intelligence Technology for Multi-speaker Dialog Modeling).

\bibliography{custom}

\end{document}

%% file: Table/main_translation.tex
\begin{table*}[t]
  \renewcommand{\arraystretch}{1.3}
  \renewcommand{\tabcolsep}{2mm}
  \centering
  \resizebox{0.8\linewidth}{!}{
  \begin{tabular}{clcccccc}
    \Xhline{3\arrayrulewidth}
    \makecell{Duration\\Controllable} & Method & ASR-BLEU $\uparrow$ & BLASER 2.0 $\uparrow$ & SIM $\uparrow$ & DNSMOS $\uparrow$ \\
    \hline
    \multirow{4}{*}{\xmark} & S2UT~\cite{lee2022direct} & 24.54 & 3.784 & 0.036 & 3.922 \\
    &CTC-S2UT~\cite{fang2024ctc} & 24.51 & 3.785 & 0.037 & 3.908 \\ 
    &UTUT~\cite{kim2024textless}$^\dagger$ & 26.49 & 3.840 & 0.036 & 3.927 \\
    &~~~w/ Zero-shot Vocoder~\cite{choi2024av2av}$^\dagger$ & 26.33 & 3.882 & 0.145 & 3.101 \\
    \cdashline{1-8}
    \multirow{5}{*}{\cmark} & TranSpeech~\cite{huang2023transpeech}$^\ddagger$ & 18.03 & - & - & - \\
    &DiffNorm~\cite{tan2024diffnorm}$^\ddagger$ & 19.53 & - & - & - \\
    &Dub-S2ST-single (Ours) & 23.88 & 3.813 & 0.036 & \textbf{3.945} \\
    &~~~w/o speed adaptation & 22.10 & 3.766 & 0.035 & 3.909 \\
    &Dub-S2ST (Ours) & \textbf{24.16} & \textbf{3.839} & \textbf{0.266} & 3.693 \\
    \Xhline{3\arrayrulewidth}
  \end{tabular}}
  \vspace{-0.1cm}
  \caption{
  Performance comparisons with state-of-the-art textless S2ST methods on CVSS-C dataset. $^\dagger$Multilingual translation model trained with a larger model size and dataset. $^\ddagger$The scores are reported from the original papers.}
  \vspace{-0.2cm}
  \label{tab:translation}
\end{table*}

%% file: Table/main_duration.tex
\begin{table}[t]
  \renewcommand{\arraystretch}{1.3}
  \renewcommand{\tabcolsep}{1mm}
  \centering
  \resizebox{0.999\linewidth}{!}{
  \begin{tabular}{lcccccccc}
    \Xhline{3\arrayrulewidth}
    Method & DC@0.2 & DC@0.4 & SC@0.2 & SC@0.4 & S. Corr \\
    \hline
    S2UT~\cite{lee2022direct} & 64.53 & 93.65 & 62.01 & 90.02 & 0.222 \\
    ~~~w/o unit deduplication & 56.45 & 89.96 & 57.88 & 85.68 & 0.254 \\
    ~~~~~~w/ pos. emb.~\cite{wu2023videodubber} & 79.67 & 99.20 & 61.76 & 88.71 & 0.355 \\
    ~~~~~~w/ pos. emb.~\cite{le2024transvip} & 80.78 & 99.16 & 61.46 & 88.61 & 0.367 \\
    CTC-S2UT~\cite{fang2024ctc} & 65.33 & 94.34 & 62.70 & 90.38 & 0.251 \\
    Dub-S2ST-single (Ours)  & \textbf{100.00} & \textbf{100.00} & \textbf{71.93} & \textbf{96.77} & \textbf{0.614} \\
    ~~~w/o speed adaptation & \textbf{100.00} & \textbf{100.00} & 66.65 & 92.16 & 0.388 \\
    \Xhline{3\arrayrulewidth}
  \end{tabular}}
  \vspace{-0.1cm}
  \caption{The performance comparisons about speech duration and speed. DC@\textit{p} and SC@\textit{p} indicate duration and speed compliance with range \textit{p}, while S. Corr denotes correlation between the syllable speed of source and generated speech.}
  \vspace{-0.1cm}
  \label{tab:duration}
\end{table}

%% file: Table/main_mos.tex
\begin{table}[t]
  \renewcommand{\arraystretch}{1.3}
  \renewcommand{\tabcolsep}{1mm}
  \centering
  \resizebox{0.999\linewidth}{!}{
  \begin{tabular}{lccc}
    \Xhline{3\arrayrulewidth}
    Method & Naturalness & \makecell{Translation\\Consistency} & \makecell{Speed\\Consistency} \\
    \hline
    S2UT~\cite{lee2022direct} & 2.60\small{ $\pm$ 0.18} & 3.32\small{ $\pm$ 0.16} & 3.62\small{ $\pm$ 0.19} \\
    CTC-S2UT~\cite{fang2024ctc} & 2.52\small{ $\pm$ 0.19} & 3.56\small{ $\pm$ 0.15} & 3.55\small{ $\pm$ 0.16} \\
    Dub-S2ST-single (Ours) & \textbf{3.37}\small{ $\pm$ 0.17} & \textbf{3.80}\small{ $\pm$ 0.13} & \textbf{3.98}\small{ $\pm$ 0.16} \\
    \Xhline{3\arrayrulewidth}
  \end{tabular}}
  \vspace{-0.1cm}
  \caption{MOS evaluation.}
  \vspace{-0.1cm}
  \label{tab:mos}
\end{table}

%% file: Table/speed_adaptation.tex
\begin{table}[t]
  \renewcommand{\arraystretch}{1.3}
  \renewcommand{\tabcolsep}{2mm}
  \centering
  \resizebox{0.77\linewidth}{!}{
  \begin{tabular}{lcc}
    \Xhline{3\arrayrulewidth}
    vs Source syllable speed & Speed adapt. & Corr \\
    \hline
    Source unit speed & & 0.606 \\
    \hdashline
    Target syllable speed & \xmark & 0.235 \\
    Target syllable speed & \cmark & \textbf{0.519} \\
    \Xhline{3\arrayrulewidth}
  \end{tabular}}
  \vspace{-0.1cm}
  \caption{Effectiveness of our speed adaptation strategy.}
  \vspace{-0.1cm}
  \label{tab:speed_adaptation}
\end{table}

%% file: Table/duration_control.tex
\begin{table}[t]
  \renewcommand{\arraystretch}{1.3}
  \renewcommand{\tabcolsep}{2mm}
  \centering
  \resizebox{0.87\linewidth}{!}{
  \begin{tabular}{lccccc}
    \Xhline{3\arrayrulewidth}
    Duration Ratio & 0.8 & 0.9 & 1.0 & 1.1 & 1.2 \\
    \hline
    Relative \# Chars & 0.851 & 0.929 & 1.000 & 1.064 & 1.133 \\
    \Xhline{3\arrayrulewidth}
  \end{tabular}}
  \vspace{-0.1cm}
  \caption{Effect of explicit duration control.}
  \vspace{-0.1cm}
  \label{tab:duration_control}
\end{table}

%% file: Table/ablation_arch.tex
\begin{table}[t]
  \renewcommand{\arraystretch}{1.3}
  \renewcommand{\tabcolsep}{2mm}
  \centering
  \resizebox{0.95\linewidth}{!}{
  \begin{tabular}{ccccccc}
    \Xhline{3\arrayrulewidth}
    \multirow{2}{*}{Decoder} & \multirow{2}{*}{Schedule} & \multicolumn{5}{c}{NFE} \\
    & & 1 & 4 & 16 & 64 & 256 \\
    \hline
    Transformer & \xmark & 2.61 & - & - & - & - \\
    Transformer & Cosine & \textbf{13.52} & 19.42 & 20.10 & 20.23 & 20.33 \\
    Transformer & Linear & 11.83 & 20.28 & 21.13 & 21.37 & 21.36 \\
    DiT & Linear & 12.09 & \textbf{20.57} & \textbf{21.65} & \textbf{22.10} & \textbf{22.27} \\
    \Xhline{3\arrayrulewidth}
  \end{tabular}}
  \vspace{-0.1cm}
  \caption{Ablation study on the model architecture and masking schedule of the speech-to-unit translation model, evaluated using ASR-BLEU.}
  \vspace{0.1cm}
  \label{tab:arch}
\end{table}

%% file: Table/ablation_loss.tex
\begin{table}[t]
  \renewcommand{\arraystretch}{1.3}
  \renewcommand{\tabcolsep}{2mm}
  \centering
  \resizebox{0.8\linewidth}{!}{
  \begin{tabular}{ccc}
    \Xhline{3\arrayrulewidth}
    Loss computation & ASR-BLEU & BLASER 2.0 \\
    \hline
    All & 21.46 & 3.742 \\
    Masked & \textbf{22.10} & \textbf{3.766} \\
    Masked (non-trivial) & 21.46 & 3.751 \\
    \Xhline{3\arrayrulewidth}
  \end{tabular}}
  \vspace{-0.1cm}
  \caption{Ablation study on loss computation during speech-to-unit translation model training.}
  \vspace{-0.1cm}
  \label{tab:loss}
\end{table}

%% file: Table/ablation_vocoder.tex
\begin{table}[t]
  \renewcommand{\arraystretch}{1.3}
  \renewcommand{\tabcolsep}{1mm}
  \centering
  \resizebox{0.999\linewidth}{!}{
  \begin{tabular}{lccc}
    \Xhline{3\arrayrulewidth}
    Model & ASR-BLEU & SIM & DNSMOS \\
    \hline
    Dub-S2ST-single & 23.88 & 0.036 & \textbf{3.945} \\
    ~~~w/ Zero-shot Vocoder~\cite{choi2024av2av} & 23.76 & 0.154 & 3.088 \\
    ~~~w/ CosyVoice VC~\cite{du2024cosyvoice} & 23.09 & \textbf{0.315} & 3.787 \\
    Dub-S2ST & \textbf{24.16} & 0.266 & 3.693 \\
    \Xhline{3\arrayrulewidth}
  \end{tabular}}
  \vspace{-0.1cm}
  \caption{Ablation study on unit-to-speech module.}
  \vspace{-0.1cm}
  \label{tab:vocoder}
\end{table}